\documentclass[runningheads]{llncs}
\usepackage[T1]{fontenc}
\usepackage{graphicx}
\usepackage[pagebackref,breaklinks,colorlinks]{hyperref}

\usepackage{color}
\usepackage{booktabs}

\usepackage{xcolor}
\usepackage{amsmath}
\usepackage{esvect}
\usepackage{comment}

\begin{document}

\title{Attend to what I say: Highlighting relevant content on slides}
%
%
\author{
Megha Mariam K M\orcidID{0009-0001-7188-9818} \and \\
C. V. Jawahar\orcidID{0000-0001-6767-7057}
}

\authorrunning{Megha and Jawahar}

\institute{
CVIT, International Institute of Information Technology, Hyderabad, India \\
\email{\{megha.km@research.,jawahar@\}iiit.ac.in} 
}

\maketitle              
\begin{abstract}

Imagine sitting in a presentation, trying to follow the speaker while simultaneously scanning the slides for relevant information. While the entire slide is visible, identifying the relevant regions can be challenging. As you focus on one part of the slide, the speaker moves on to a new sentence, leaving you scrambling to catch up visually. This constant back-and-forth creates a disconnect between what’s being said and the most important visual elements, making it hard to absorb key details—especially in fast-paced or content-heavy presentations, as in a conference talk. This will require an understanding of slides, especially text, graphics, and layout. We introduce a method that automatically identifies and highlights the most relevant slide regions based on the speaker’s narrative. By analyzing spoken content and matching it with textual or graphical elements in the slides, our approach ensures better synchronization between what you hear and what you need to attend to. We explore different ways of solving this problem and assess their success and failure cases. Analyzing multimedia documents is emerging as a key requirement for seamless understanding of content-rich videos, such as educational videos and conference talks, by reducing cognitive strain and improving comprehension. Code and dataset available at: \url{https://github.com/meghamariamkm2002/Slide_Highlight}

\end{abstract}
\begin{figure}[!t]
\centering
\includegraphics[height=0.5\textwidth,width=1.01\textwidth]{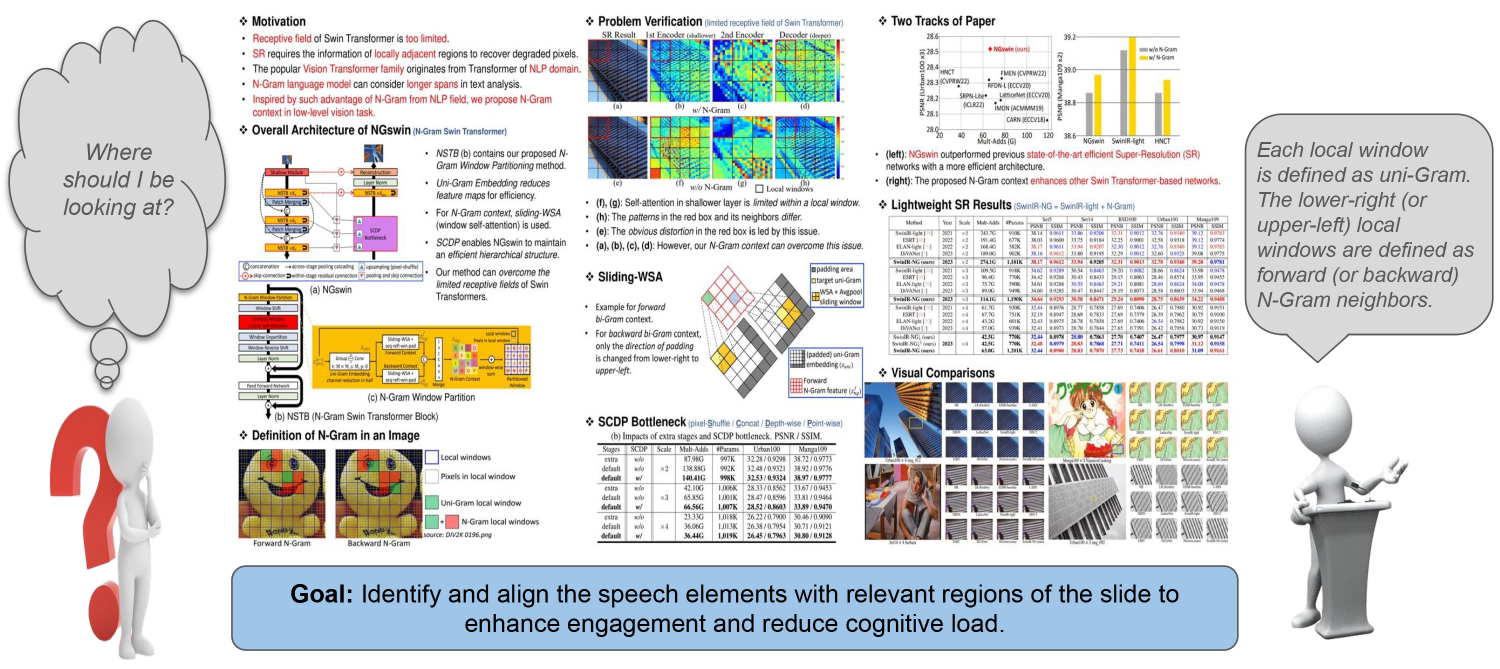}
\caption{This illustration highlights the challenge audiences face during presentations in simultaneously locating relevant content on slides while paying attention to the speaker. By synchronizing spoken information with corresponding visual elements, cognitive load is reduced, engagement is improved, and comprehension and retention of key insights are enhanced.}
\label{fig:informal-prob-defn}
\end{figure}
\section{Introduction}
Attending a multimedia presentation often feels like a balancing act—listening to the speaker while simultaneously scanning the slides for relevant information. The entire slide is available, yet identifying the most critical regions can be challenging. As the speaker moves through different aspects of the content, audience members must quickly locate the corresponding visual elements to fully grasp the explanation. However, spoken language is fleeting—by the time they find the right section, the discussion has already moved forward. This misalignment forces attendees to split their attention between processing speech and searching for visual cues, increasing cognitive effort and making it hard to absorb information effectively. Over time, this struggle can lead to information loss, reduced engagement, and a diminished overall learning experience. Figure~\ref{fig:informal-prob-defn} illustrates this challenge.

How do we address this problem? One simple strategy should be to highlight the relevant regions. However, this is not trivial as we need to work together with speech and slides. When speakers reference multiple slide elements at once or fail to visually emphasize key content in sync with their narration, the difficulty becomes even greater. This lack of coordination places an additional cognitive burden on the audience, requiring them to process auditory information while searching for relevant visuals—ultimately impacting comprehension and engagement. Research on cognitive load in multimedia learning suggests that an increased working memory demand can negatively affect learning outcomes~\cite{Anmarkrud03042019}. Furthermore, individual differences in how people process multimedia content play a crucial role—when the presentation format does not align with their preferred learning style, the struggle intensifies~\cite{Cog2}.

Solutions to problems like this require comprehensive solutions for multi-modal document understanding. At this stage, the modalities are not aligned at a fine-grained level. There have been many attempts in the past to align modalities such as movies and scripts, speech and transcripts, and books and texts~\cite{books_and_movies_yukun,align_sub_sign_lang}. In this case, we need to align spoken content with relevant slide elements to create a seamless and intuitive presentation experience. To validate the method, we introduce a dataset of conference presentations. The dataset contains video segments with speech annotated with corresponding speech transcripts with timestamps, slide layouts, and the text presented on the slide. We explore multiple methods for aligning the speaker’s narrative with both textual and visual components of the slides, ensuring that key information is effortlessly accessible. We hope that research on problems like this will help in a better understanding of multi-modal documents, leading to many potential applications in the real world.

\section{Related Work}
\vspace{-0.05in}

\subsection{Human Document Interaction}
We as humans increasingly interact with documents today; we have gone well beyond "reading" books and historical manuscripts. We "watch" videos, such as educational videos, and learn concepts rather than "reading" books and handwritten notes as in the traditional educational system~\cite{can_videos}. Such visual documents are rich in content, speech, text, graphics, and annotations. From a document understanding point of view, one might want to pass the visual stream through the layout and textual content. Extensive research in \textsc{ocr}, with appropriate adaptation, has been increasingly used. Speech can also be converted to text with timestamps using Automatic Speech Recognition (\textsc{asr}) tools~\cite{radford2022robustspeechrecognitionlargescale,bain2023whisperxtimeaccuratespeechtranscription}. In recent years, methods like Whisper have provided high speech-recognition capability in all languages. However, combining these two modalities is still challenging. Audio-visual processing of videos has been receiving attention in recent years~\cite{shahabaz2024increasing,singh2023unsupervised,1021881,8962278}.

Though the accuracies of speech recognition have been rapidly improving with better and better performance, technical presentations could pose newer challenges to commercial \textsc{asr} systems, as the script could contain technical terms that are unique to a domain. A typical international conference today also sees diverse accents, including those that commercial systems are not well trained for. This is also true for \textsc{ocr} systems that can be used for parsing slides. Though \textsc{ocr}s and \textsc{asr}s have been commercially available with good accuracies, there is a need to improve their accuracies with domain-specific techniques such as post-processing. In the past, researchers have used lexicons and \textsc{llm}s for post-processing recognition outputs~\cite{kumari2022lexiconattentionbasedhandwritten,kumari2023comprehensivehandwrittenparagraphtext,lakomkin2023endtoendspeechrecognitioncontextualization}.

\subsection{Understanding Slide Documents}
Presentations are essential for education and information sharing, with slides complementing the talk. Research has focused on analyzing slides to improve their structure and accessibility. Competitions such as the Page Segmentation Competition~\cite{Page-Seg} and \textsc{icdar} 2017 \textsc{rdcl}~\cite{ICDAR-2017-Doc-Anal-Recog} have set benchmarks for layout analysis, while the \textsc{icdar} 2023 competition on structured text extraction~\cite{icdar-2023-compet} emphasized the challenge of extracting meaningful text from complex documents, particularly under zero-shot and few-shot conditions. A key aspect of slide analysis is layout understanding, which involves identifying distinct regions like text, images, and charts while maintaining semantic coherence. Deep learning-based approaches have significantly advanced this field. Yang and Hsu (2020)~\cite{Yang-2021} introduced \textsc{slld}, a \textsc{cnn}-based model that approaches segmentation as an object recognition task, outperforming traditional methods. Expanding on this, Yang and Hsu (2022)~\cite{Yang-2022} developed \textsc{trdlu}, a transformer-based model that classifies slide content more accurately using an encoder-decoder framework. Beyond digital slides, real-world applications have been explored. WiSe~\cite{WiSE} enables segmentation from live slide photos, allowing structured information extraction even in dynamic settings such as conferences and classrooms. Information retrieval techniques further enhance slide accessibility. \textsc{ocr}-based slide retrieval~\cite{ocr-slide-ret} achieves accuracy levels comparable to traditional electronic document search. Additionally, video-based frameworks dynamically analyze slide content. The SlideVQA dataset~\cite{SlideVQA}, consisting of over 2,600 slide decks and 52,000 images, supports complex reasoning tasks. A document \textsc{vqa} model integrating evidence selection and question answering has also surpassed previous approaches, though it still lags behind human comprehension. With advancements in deep learning, multimodal processing, and enriched datasets, research continues to refine slide analysis techniques, bringing us closer to machine-level document understanding.

\subsection{Alignment in Document Understanding}
\vspace{0.05in}
Aligning different modalities has been a fundamental tool in various document understanding tasks. One successful application is word-spotting, where text is matched with word images using Dynamic Time Warping (\textsc{dtw}) and Hidden Markov Models (\textsc{hmm})~\cite{Word-img-matching}. Alignments have also been used to create annotated data by linking document images with paragraphs and pages of text. Such alignments enable scalable and cost-effective annotation of words, facilitating the rapid development of recognition algorithms with minimal human supervision.

It is very common today to align videos with transcripts to create temporal annotations, aiding in retrieval and weakly supervised captioning~\cite{align_sub_sign_lang}. For example, books are often aligned with movies~\cite{books_and_movies_yukun}. Fine-grained alignment can help in many challenging annotation tasks, such as phoneme recognition and sign language translation \cite{align_sub_sign_lang}.

Matching is also carried out across languages and modalities. Matching is also tried with the help of semantic similarities. For example, a word in one language could be matched (or aligned) with i) a word written in a different font or by a different person, ii) a synonym of the word appearing in print or handwritten form, iii) an equivalent word appearing in a different language, iv) the word appearing in speech, v) the word or associated object appearing in a video stream.

\section{Methodology}
This work presents a framework for aligning spoken content with presentation slides to enhance audience engagement. By establishing a clear correspondence between speech and visuals, the approach promotes better comprehension and reduces cognitive load during presentations. Below, we describe our method in detail.

\subsection{Problem Definition}  
\begin{figure}
    \centering
\includegraphics[height=0.5\textwidth,width=1\textwidth]{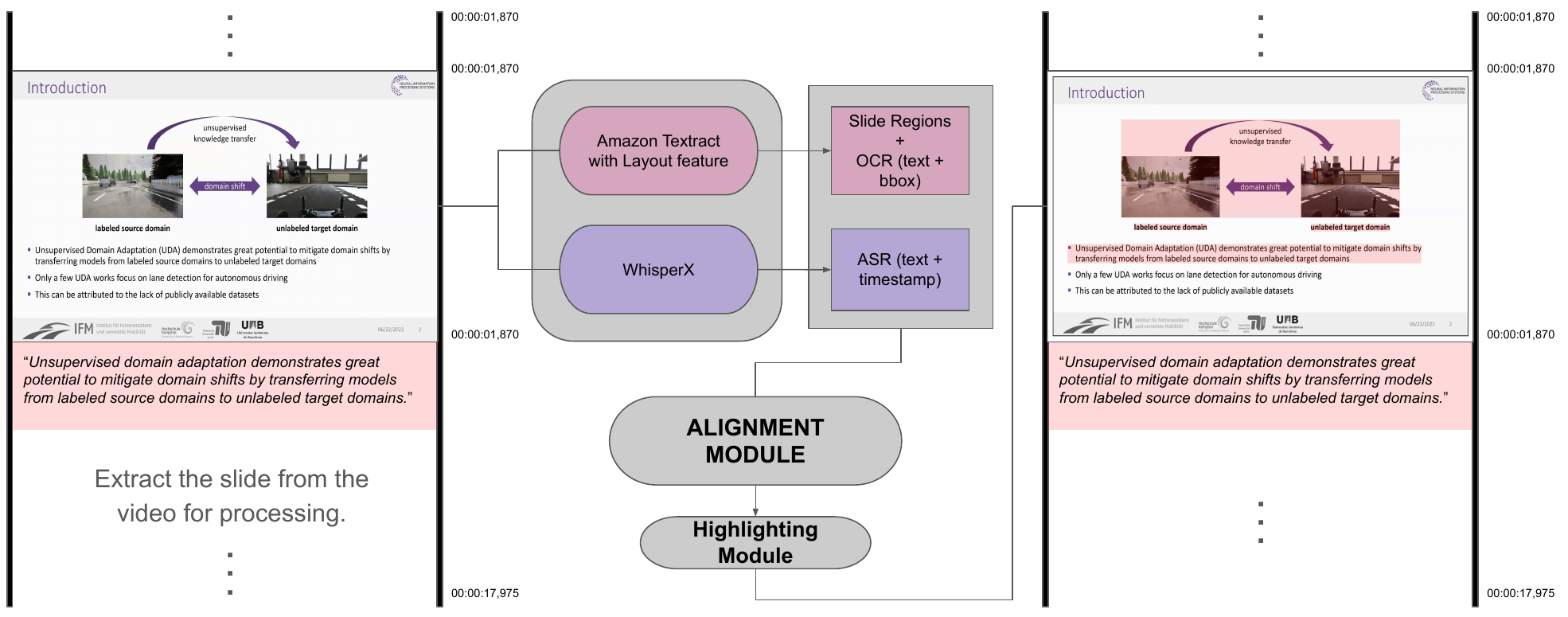}
    \caption{This figure illustrates the pipeline of our method. The Alignment Module takes \textsc{ocr}-extracted text and \textsc{asr}-generated transcripts as inputs to establish correspondences between spoken and visual content. These aligned results are then passed to the Highlighting Module, from which a suitable visualization can be chosen to highlight the relevant slide content.}
    \label{fig:formal-prob-defn}
\end{figure}

Given a presentation slide and its corresponding speech, the objective is to establish a correspondence between the spoken content and specific regions of the slide. This involves mapping the speech to relevant areas within the slide. As illustrated in Figure~\ref{fig:formal-prob-defn}, the alignment module utilizes the outputs of \textsc{ocr} and \textsc{asr} for alignment. Once the relevant slide regions are identified, the results are passed to the highlighting module, from which a specific visualization can be selected for presentation.

\vspace{-0.3in}
\subsection{Dataset}
\begin{figure}
\centering
\includegraphics[height=0.55\textwidth,width=1.0\textwidth]{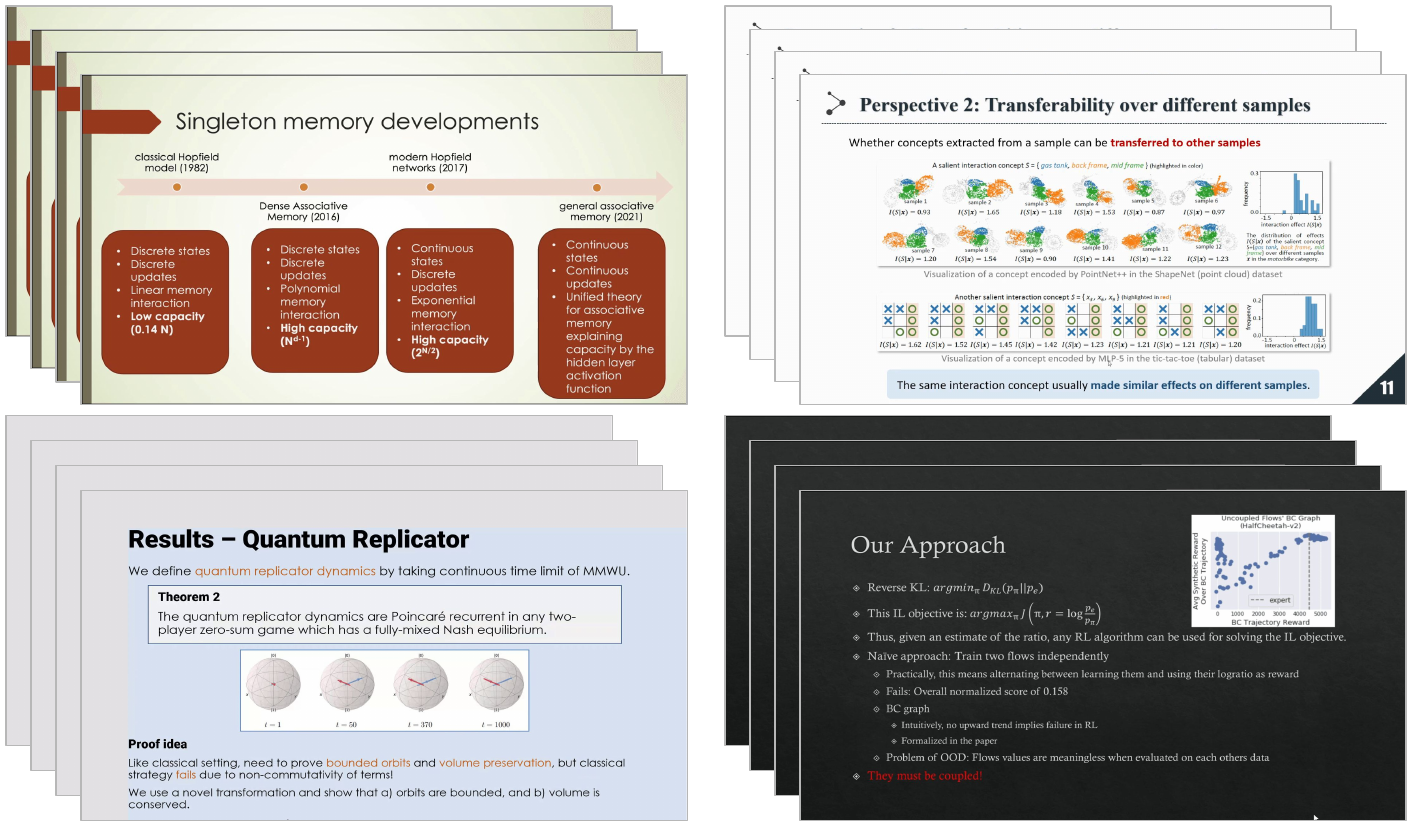}
\caption{This figure showcases a set of presentation slides from our dataset. The dataset consists of slides with varying layouts, color schemes, and content structures, reflecting the diversity of real-world academic and professional presentations.}
\label{fig:dataset}
\end{figure}

\paragraph{\textbf{\textit{Data Collection:}}}
We curated a dataset comprising 14 presentation videos from NeurIPS 2022 and ICML 2023 conference presentations available on the SlidesLive platform. Each slide is paired with its corresponding audio segment and transcript. The dataset consists of 150 slides, encompassing diverse visual and textual elements such as figures, equations, tables, and text in various fonts and sizes. Figure~\ref{fig:dataset} presents a selection of representative samples from our dataset, illustrating the diversity in both visual and textual content.

\begin{figure}[!htbp]
    \centering
    \begin{minipage}{0.4\textwidth}
        \centering
        \begin{tabular}{cccccc}
            \toprule
            & Min & Max & Avg & Median \\
            \midrule
            Duration (in sec) & 2 & 56 & 21.10 & 18 \\
            Word Count in \textsc{ocr} & 1 & 799 & 99.93 & 83 \\
            Word Count in \textsc{asr} & 5 & 307 & 72.71 & 59 \\
            \bottomrule
        \end{tabular}
        \caption{Statistics for Audio Segment Durations, Word Count in \textsc{asr} and \textsc{ocr}}
        \label{tab:audio_stats}
    \end{minipage}%
    \hfill
    \begin{minipage}{0.45\textwidth}
        \centering
        \includegraphics[height=0.55\textwidth, width=\textwidth]{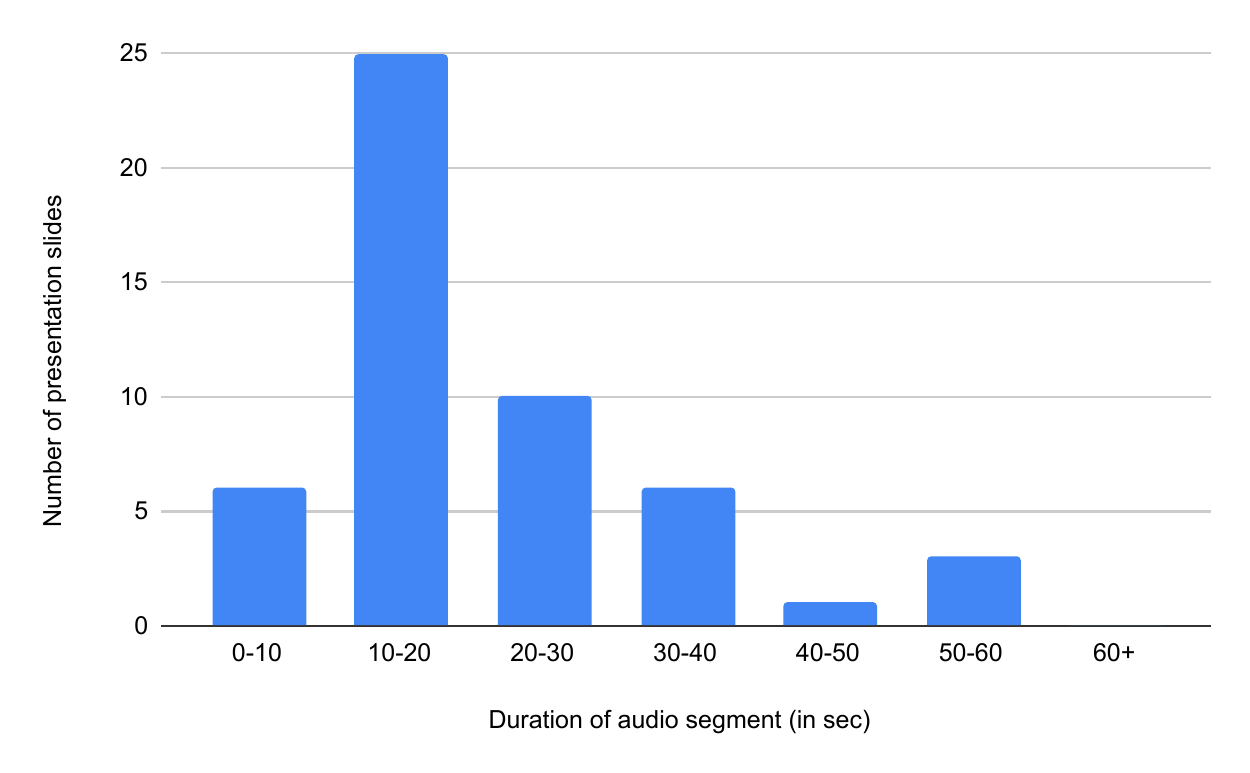}
        \caption{The distribution of presentation slides based on the duration of their corresponding audio segments.}
        \label{fig:distribution}
    \end{minipage}
\end{figure}

\paragraph{\textbf{\textit{Statistics:}}}
The total duration of the dataset exceeds one hour. Figure~\ref{fig:distribution} illustrates the distribution of presentation slides based on the duration of their corresponding audio segments, highlighting that most slides are associated with audio segments lasting between 10 and 20 seconds. The slide text corpus contains 7,466 unique alphabetical words, while the transcript text comprises 6,708 unique alphabetical words. Additionally, Figure~\ref{tab:audio_stats} provides statistics on the audio segment durations, word count in \textsc{ocr}, and word count in \textsc{asr}, including the minimum, maximum, average, and median values. This dataset serves as a valuable resource for evaluating the approach discussed below.

\vspace{-0.25in}
\subsection{Recognizing text in slides and speech}
\begin{figure}[!h]
\centering
\includegraphics[height=0.8\textwidth,width=1\textwidth]{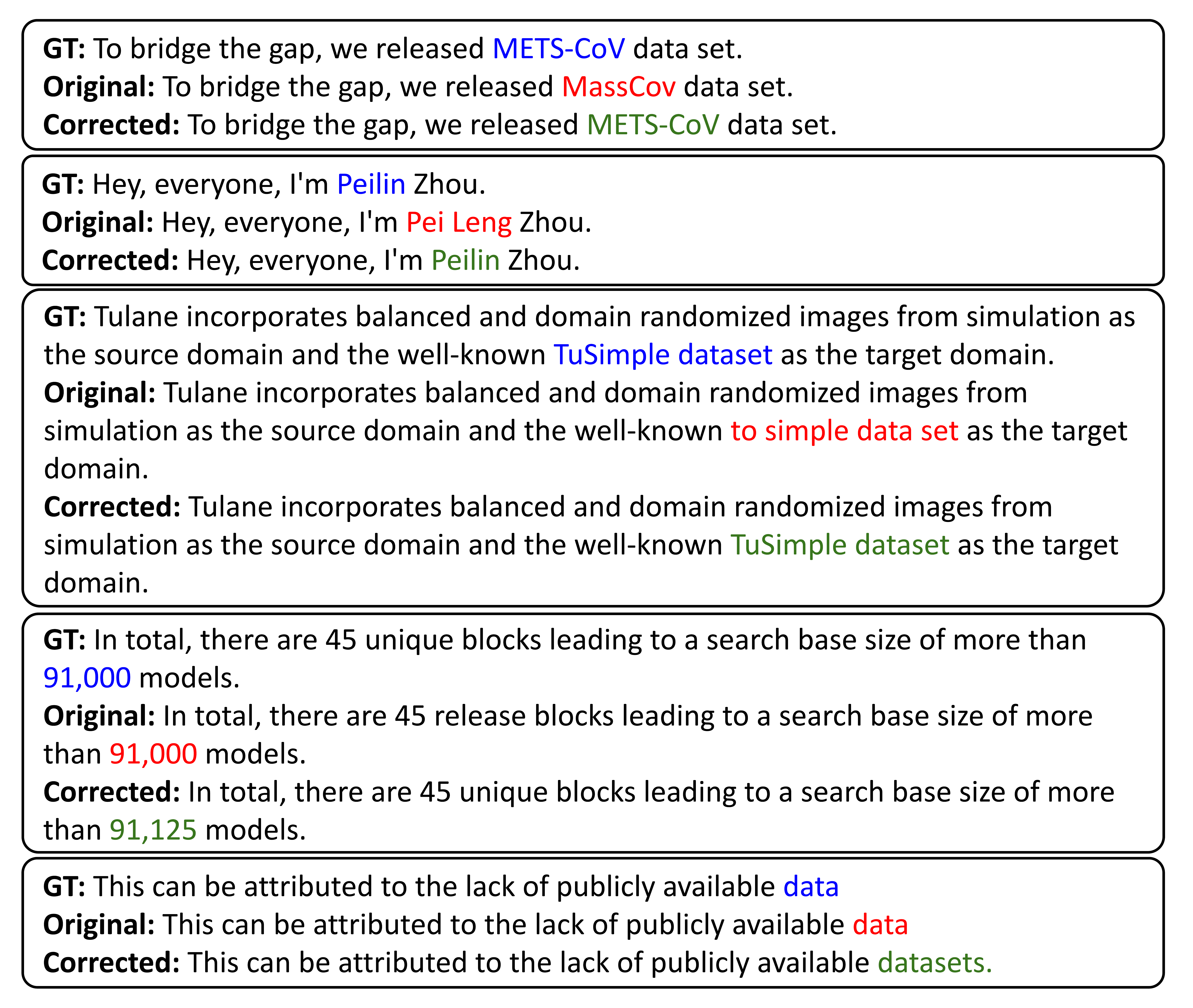}
\caption{Comparison of \textsc{asr}-generated text (\textsc{gt}) with the original and \textsc{ocr}-guided corrected versions.}
\label{fig:ASR-Post-Cor}
\end{figure}

\begin{figure}[!htbp]
\centering
\includegraphics[height=0.6\textwidth,width=1\textwidth]{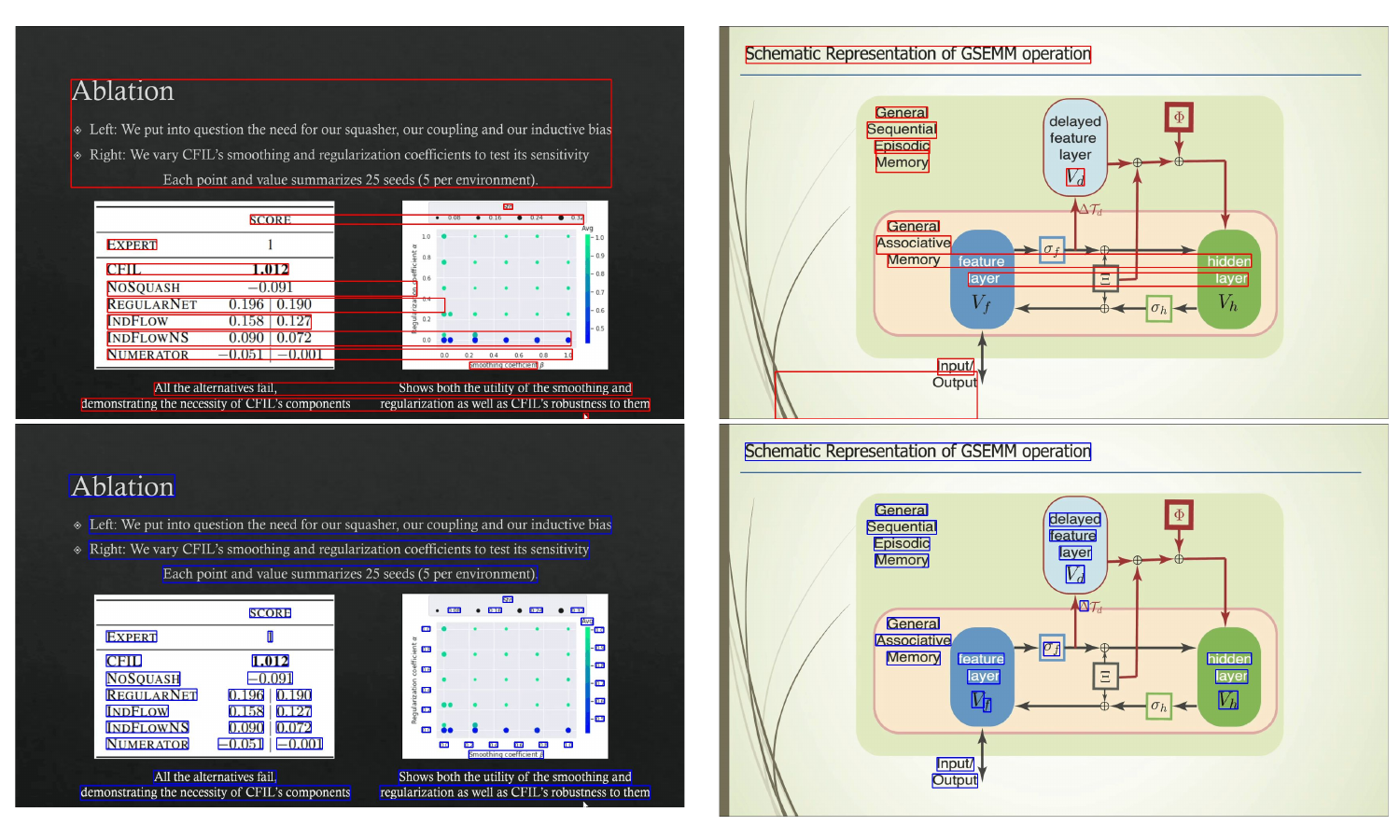}
\caption{Comparison of \textsc{ocr} results from Tesseract (red) and \textsc{aws} Textract (blue) on slides with complex layouts.}
\label{fig:ocr-comp}
\end{figure}

Presentation slides often contain structured layouts with tables and dense text, making accurate segmentation critical for aligning visual and spoken content. We use Amazon Textract\footnote{\url{https://aws.amazon.com/textract/}} for its OCR and layout analysis capabilities, enabling precise segmentation into meaningful regions. For speech, we employ WhisperX \cite{bain2023whisperxtimeaccuratespeechtranscription}, which achieves low error rates (\textsc{cer} = 0.0129, \textsc{wer} = 0.0297; see Table~\ref{tab:ASR-Results}), though it struggles with technical terms, speaker names, and accents (Figure~\ref{fig:ASR-Post-Cor}). While Textract handles structured content well, open-source OCR tools like Tesseract often fail on multi-line text, complex fonts, or embedded figures (Figure~\ref{fig:ocr-comp}), underscoring the need for advanced OCR like Textract.

\begin{table}[!htbp]
    \centering
    \begin{tabular}{l|cccccc}
    \toprule
     & \textsc{cer} & \textsc{wer} & Edit Distance & Precision & Recall & F1-score \\
    \midrule
    WhisperX  & 0.0129 & 0.0297 & 1.0327 & 0.9745 & 0.9718 & 0.9731 \\
    Corrected & 0.0366 & 0.0832 & 3.3224 & 0.9287 & 0.9266 & 0.9271 \\
    \bottomrule
\end{tabular}
    \vspace{.05in}
    \caption{Comparison of \textsc{asr} performance before and after post-correction using \textsc{ocr}. Metrics include Character Error Rate (\textsc{cer}), Word Error Rate (\textsc{wer}), Edit Distance, Precision, Recall, and \textsc{F1-score}. WhisperX represents the original \textsc{asr} output, while "Corrected" refers to the \textsc{asr} output refined using \textsc{ocr}-based post-correction.}
    \label{tab:ASR-Results}
\end{table}

\vspace{-0.4in}
\subsection{Alignment of speech and text}

\begin{figure}[!htbp]
\centering
\includegraphics[height=0.35\textwidth]{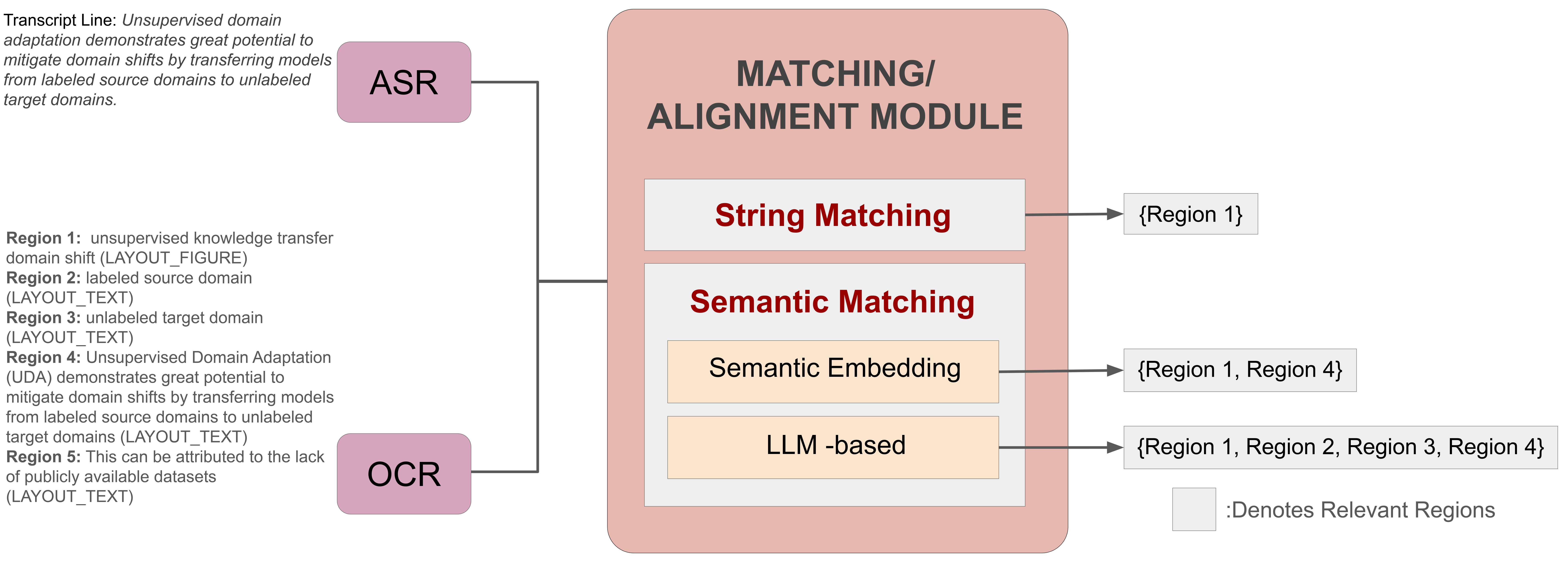}
\caption{The figure illustrates the different categories in the matching module and how they function for an given example.}
\label{fig:alignment}
\end{figure}

Our approach matches \textsc{ocr}-extracted text from slides with \textsc{asr}-generated transcripts to identify relevant slide regions corresponding to spoken content, as illustrated in Fig.~\ref{fig:alignment}. There are two broader methods for matching: (1) String Matching: which identifies exact or near-exact textual correspondences, and (2) Semantic Matching: which leverages contextual similarity to establish meaningful associations between spoken content and visual elements. We will discuss these methods in more detail below.

\paragraph{String Matching}
One straightforward method for matching \textsc{ocr} and \textsc{asr} outputs is fuzzy matching. This approach relies on identifying textual similarities, even in the presence of minor differences such as spelling errors, word order variations, or formatting inconsistencies between the spoken content and the slide text. While effective for simple matches, the accuracy of this method may decrease in more complex scenarios where the alignment is less direct.

\subsubsection{Semantic Alignment}
\paragraph{\textbf{Semantic Embeddings: }}
A more advanced approach is to use semantic similarity for aligning transcript lines to slide regions. This method compares the semantic embeddings of the transcript lines with the \textsc{ocr}-generated text of slide regions. Unlike fuzzy matching, which relies on surface-level text similarity, semantic similarity captures the underlying meaning of the content, allowing for more robust alignment even in the presence of paraphrasing or significant lexical variation. To achieve this, we employ a variety of pre-trained models that generate semantic embeddings.
\newline
\textsc{Mini\textsc{lm}\cite{MiniLM}:} A lightweight, efficient model known for producing high-quality sentence embeddings that capture the semantic meaning of the text.
\newline
\textsc{Sci-\textsc{bert} \cite{sci-bert}:} A specialized \textsc{bert}-based model fine-tuned on scientific literature, making it well-suited for technical and domain-specific content often found in academic and professional presentations.
\newline
\textsc{specter \cite{specter}:} A model trained on scholarly documents to generate embeddings that are particularly effective in capturing the relationships and semantic structures within academic and scientific text.

\paragraph{\textbf{\textsc{llm}: }}
\textsc{llm}s such as Flan-T5 and Qwen-Instruct 2.5 are utilized for further enhancing the alignment process, in addition to above mentioned semantic embedding models. 
\newline
Flan-\textsc{t5} \cite{FlanT5}: An instruction-tuned version of the \textsc{t5} model (Text-to-Text Transfer Transformer), Flan-\textsc{t5} excels at following specific instructions, making it highly suitable for post-processing tasks. We go through each ocr region and ask Flan-\textsc{t5} whether the ocr region is relevant to the transcript line as a yes/no question for a given transcript line.
\newline
Qwen-Instruct 2.5 \cite{Qwen2.5}: A powerful language model that excels in instruction-following tasks. In this approach, the model is presented with a transcript line along with the \textsc{ocr} outputs from all slide regions, and its task is to select the most relevant slide regions for the given transcript line.

\paragraph{}
For our dataset, the best thresholds were: Fuzzy (textual: 0.45), S-BERT (textual: 0.55), SPECTER (textual: 0.85), and Sci-BERT (textual: 0.75), each achieving the highest F1-score for its respective method. Since optimal thresholds may not be known beforehand, we used three general-purpose threshold settings—T-1 (textual: 0.8, visual: 0.6), T-2 (textual: 0.7, visual: 0.6), and T-3 (textual: 0.6, visual: 0.6)

\subsection{Highlighting Regions}
With the correspondence between the speech and slide regions established, the next step is to visualize the relevant slide regions in alignment with the spoken narrative. Several approaches can be used, such as drawing a box around the region, shading it with a distinct color, magnifying or enlarging the area of interest, or completely removing non-essential slide content. If the goal is to allow users to follow the speech while focusing on the highlighted regions but still explore other parts of the slide, then hiding the background content may not be a suitable choice. However, if the slide contains multiple distinct sections covering different topics that are not related to each other, removing irrelevant regions may be a more effective approach to maintaining clarity and focus.

\subsection{Using \textsc{ocr} to improve \textsc{asr} accuracy}
\textsc{asr} in presentations often misinterprets technical or domain-specific terms, affecting alignment accuracy. To address this, we employ a prompt-based \textsc{llm} approach that leverages slide context to refine \textsc{asr} output. We use Qwen/Qwen2.5-1.5B-Instruct2~\cite{Qwen2.5} with both transcription and slide text as input to correct misspellings and improve recognition of key terms. As shown in Fig.~6, this enhances the accuracy of domain-specific content identification.

\begin{figure}
    \centering
    \includegraphics[height=0.55\textwidth,width=1.0\textwidth]{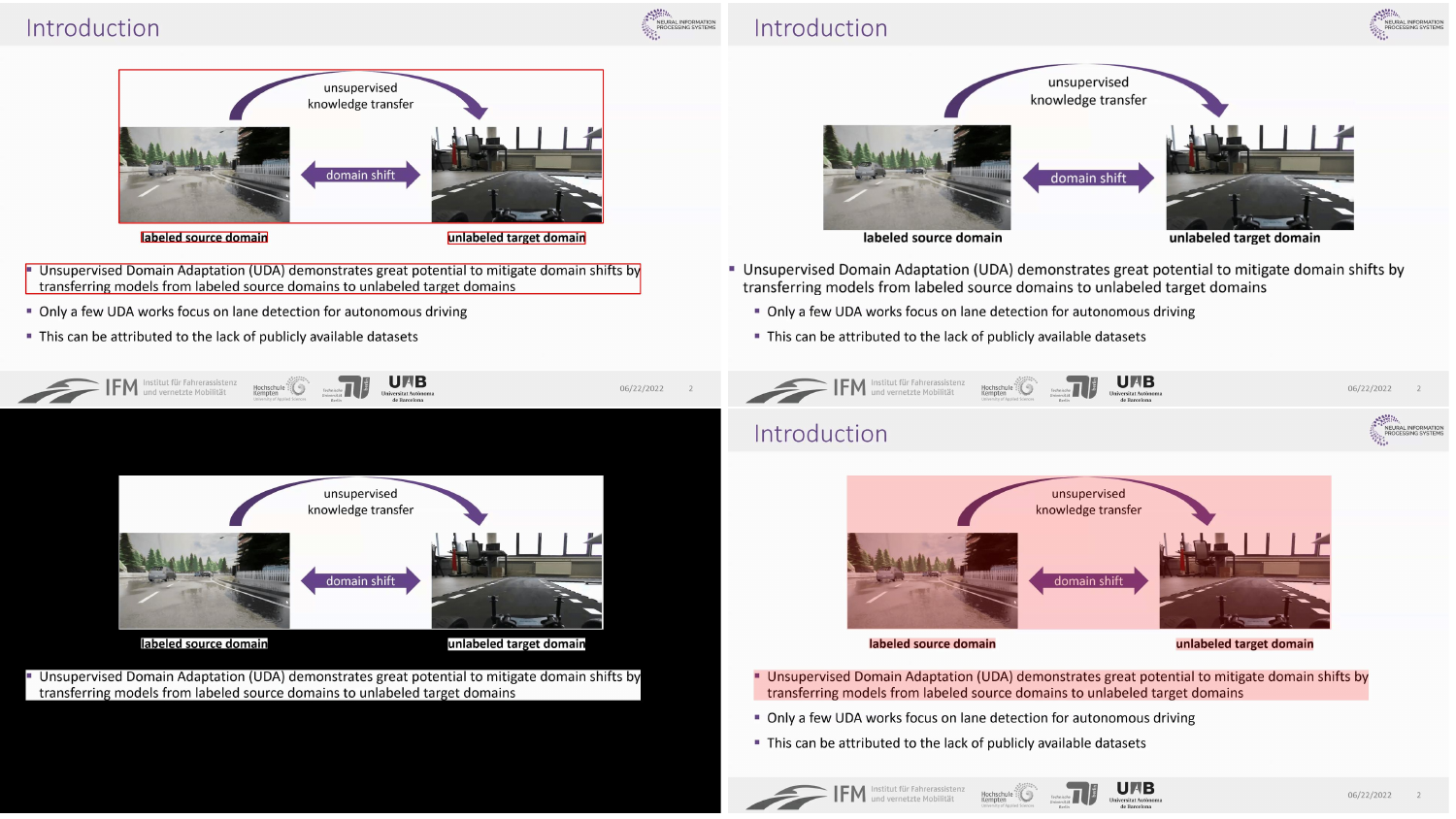}
    \caption{Illustration of different techniques for highlighting relevant regions in presentation slides, including transparent bounding boxes, color overlays, content removal, and magnification of key areas.}
    \label{fig:enter-label}
\end{figure}

\section{Evaluation}
\label{sec:Evaluation}
\subsection{Metric}  

\paragraph{\textbf{Correctness Score:}}  
The Correctness Score (\( S_c \)) evaluates how accurately the identified slide regions match the corresponding transcript content. A high score indicates that most detected regions align correctly with the transcript, while a low score suggests errors, such as incorrect regions being identified or excessive regions being detected for a given transcript line. If no predictions are made for a transcript line, the score is automatically set to 1. The Incorrectness Score, defined as \( 1 - S_c \), represents the proportion of incorrect alignments. 

\paragraph{\textbf{Missing Score:}}  
The Missing Score (\( S_m \)) measures how many relevant slide regions were overlooked during alignment. A high missing score suggests that relevant regions were not detected, leading to incomplete alignments. It is the ratio of missing alignments in the prediction to the total number of expected alignments for a transcript line. If no alignments are needed for a transcript line, the score is set to 0.

\paragraph{\textbf{F1-Score: }} 
To compute the F1-Score, both \textit{precision} and \textit{recall} must be defined. Recall is given by \(1 - S_m\). Precision is defined as the proportion of correctly predicted regions among all predicted regions. It is important to note that precision is \textbf{not} equivalent to the Correctness Score. If no predictions are made for a given transcript line, the precision is defined as zero.

\begin{figure}[!htbp]
\centering
\includegraphics[height=0.7\textwidth,width=1.0\textwidth]{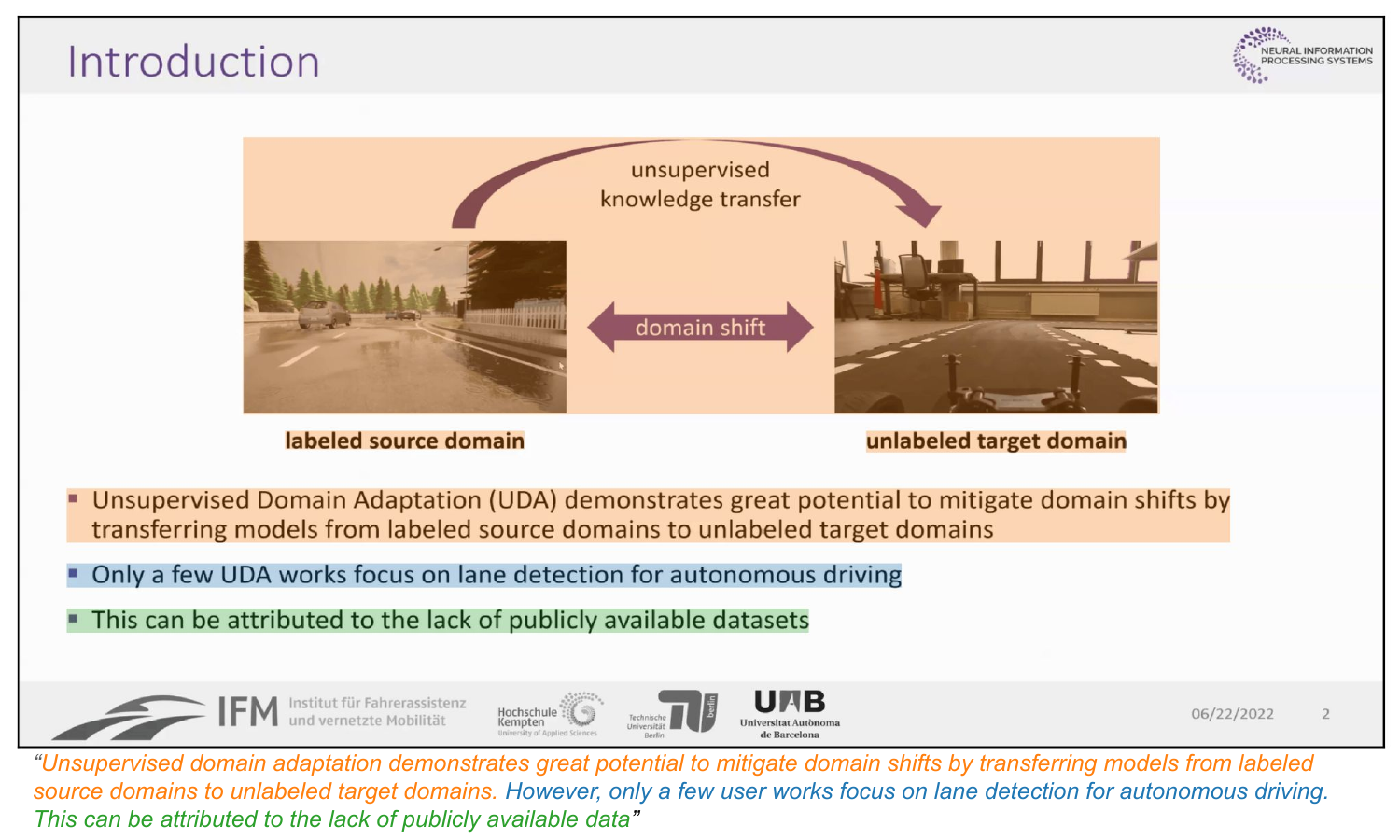}
\caption{The figure illustrates the method’s output, with each transcript line and its corresponding slide region highlighted in the same color.}
\label{fig:qualitative-1}
\end{figure}
\begin{figure}
\centering
\includegraphics[height=0.6\textwidth,width=1\textwidth]{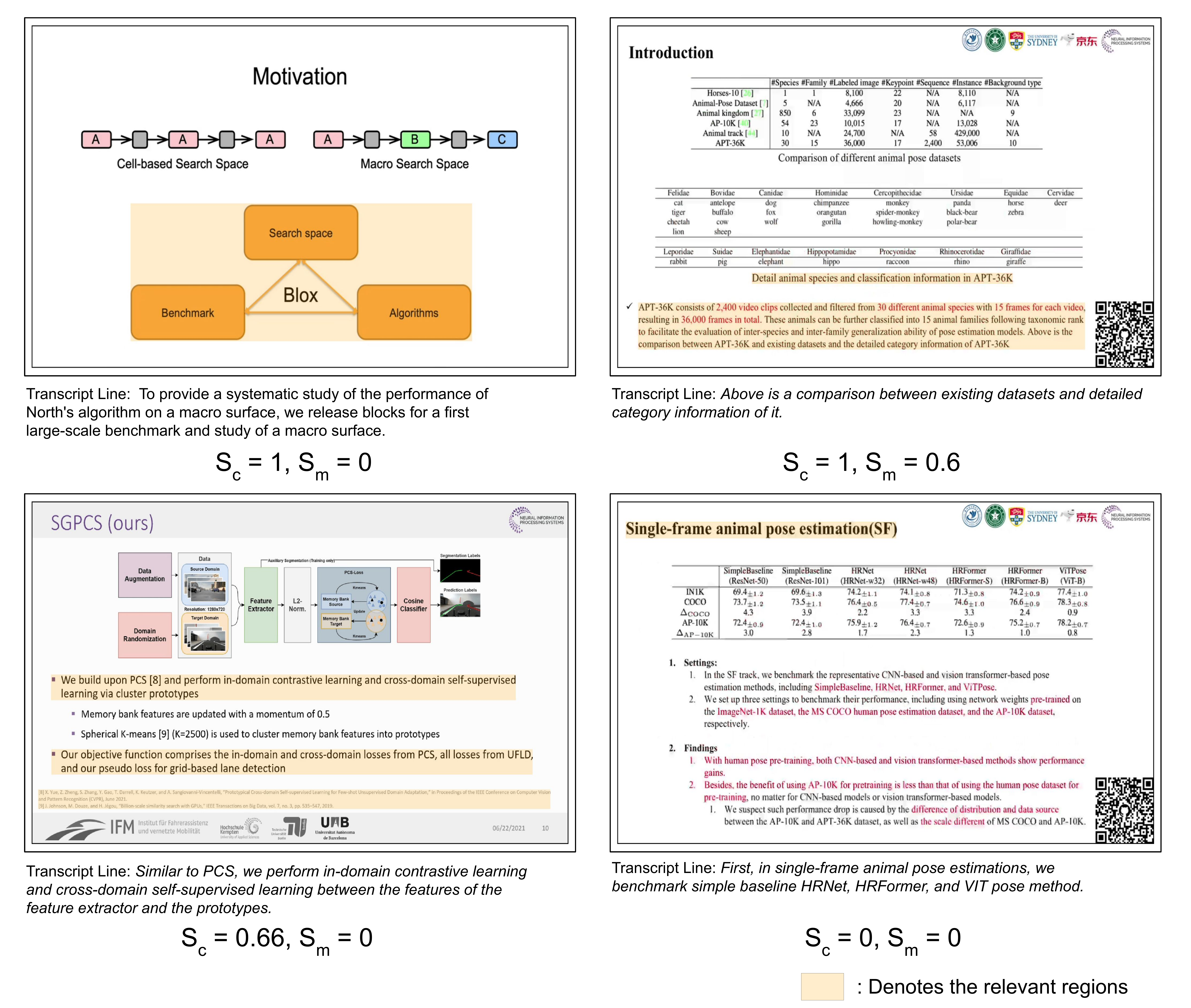}
\caption{\text{t5}-based alignment results across three examples: the first shows perfect correctness and missing scores (\( S_c = 1, S_m = 0 \)); the second has perfect correctness (\( S_c = 1 \)); and the third has perfect missing score (\( S_m = 0 \)).
}
\label{fig:qualitative-2}
\end{figure}

\subsection{Results}
\subsubsection{\textsc{ocr} and \textsc{asr} Matching Results}
\begin{figure}[!htbp]
\centering
\includegraphics[height=0.5\textwidth, width=1\columnwidth]{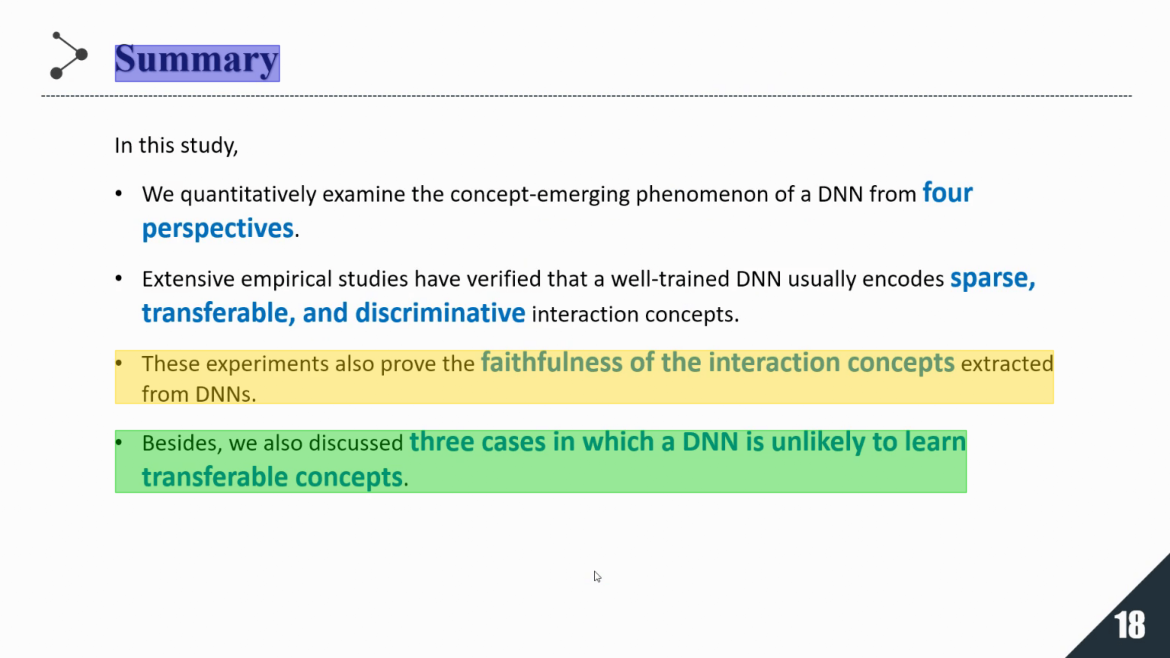}
\caption{The figure shows that for "That's all," \textsc{t5} highlights blue and Qwen highlights yellow; for "Thank you for your attention," \textsc{t5} highlights blue and Qwen highlights green.} 
\label{fig:failure_case}
\end{figure}

\paragraph{\textbf{Qualitative Results: }}
Semantic alignment demonstrates good performance in mapping spoken content to visual slide regions. As shown in Figure~\ref{fig:qualitative-1}, \textsc{t5} effectively links transcript lines to their corresponding areas on the slide. To better understand the behavior of alignment models, Figure~\ref{fig:qualitative-2} presents an analysis of Flan-\textsc{t5} across four representative cases: (1) perfect alignment with \( S_c = 1 \), \( S_m = 0 \); (2) perfect correctness with additional mismatches (\( S_c = 1 \), \( S_m = 0.6 \)); (3) perfect missing score with partial correctness (\( S_c = 0.66 \), \( S_m = 0 \)); and (4) a failure case where predictions are made despite no true match (\( S_c = 1 \), \( S_m = 0 \)). Interestingly, S-BERT T-3 achieves the best F1 score overall. This performance arises from the fact that while LLMs like \textsc{t5} and Qwen are context-aware and can handle complex semantic matches, they sometimes misinterpret generic phrases, which leads to incorrect region predictions. For example, Qwen associates phrases like ``That’s all'' or ``Thank you for your attention'' with conclusion sections, while \textsc{t5} aligns them with the ``Summary'' label, as illustrated in Figure~\ref{fig:failure_case}. On the other hand, simpler baselines like Fuzzy rely on near-exact string matches and tend to make fewer predictions. This conservative behavior results in inflated correctness scores (e.g., \( S_c = 1 \)) when no match is predicted, even if relevant matches are missed. The missing score component helps account for such omissions, providing a more balanced evaluation of alignment quality.

\paragraph{\textbf{Quantitative Results}}
The effectiveness of the proposed alignment method is evaluated using the Correctness Score (\( S_c \)) and Missing Score (\( S_m \)), as presented in Table~\ref{tab:method_scores}. Fuzzy matching consistently achieves the highest correctness (\( S_c \approx 0.99 \)); however, it suffers from high missing scores, particularly at T-1 (\( S_m = 0.663 \)), due to its inability to detect more difficult matches. In contrast, embedding-based methods such as Sci-\textsc{bert} and \textsc{specter} perform poorly overall, with correctness scores decreasing sharply across thresholds. Among the evaluated methods, Mini\textsc{lm} T-1 yields the best F1 score, followed closely by \textsc{t5}. \textsc{llm}-based models offer a better balance between correctness and missing: Flan-\textsc{t5} achieves \( S_c = 0.687 \) and \( S_m = 0.272 \), while Qwen-2.5 shows higher correctness (\( S_c = 0.843 \)) at the cost of a higher missing score (\( S_m = 0.490 \)). When comparing alignments on original (\textsc{asr}) versus corrected transcripts, \textsc{llm}s like Qwen benefit in correctness (e.g., a gain of +0.084 in \( S_c \)), and most methods exhibit improved missing scores. Overall, while fuzzy matching excels at exact match scenarios due to its high correctness, \textsc{llm}s demonstrate more robust semantic alignment capabilities, particularly when provided with corrected input.

\begin{table}[h]
    \centering
    \renewcommand{\arraystretch}{1.2}
    \begin{tabular}{lcccccc}
    \toprule
    \textbf{Method} &   \textbf{Avg \( S_c \)} & \textbf{Avg \( S_m \)} & \textbf{Avg \( F_1 \)} & \textbf{Avg \( S_c \)} (W) & \textbf{Avg \( S_m \)} (W) & \textbf{Avg \( F_1 \)} (W) \\
    \midrule
    Fuzzy [T-1] & 0.998 & 0.663 & 0.254 & 1.000 & 0.673 & 0.246 \\
    Fuzzy [T-2] & 0.998 & 0.603 & 0.307 & 1.000 & 0.617 & 0.295 \\
    Fuzzy [T-3] & 0.989 & 0.525 & 0.364 & 0.989 & 0.526 & 0.360 \\
    \textsc{s-bert} [T-1] & 0.983 & 0.578 & 0.335 & 0.993 & 0.624 & 0.299 \\
    \textsc{s-bert} [T-2] & 0.943 & 0.460 & 0.403 & 0.966 & 0.507 & 0.373 \\
    \textsc{s-bert} [T-3] & 0.868 & 0.372 & 0.429 & 0.902 & 0.391 & 0.430 \\
    Sci-\textsc{bert} [T-1] & 0.524 & 0.356 & 0.280 & 0.636 & 0.474 & 0.279 \\
    Sci-\textsc{bert} [T-2] & 0.313 & 0.169 & 0.254 & 0.307 & 0.169 & 0.252 \\
    Sci-\textsc{bert} [T-3] & 0.222 & 0.056 & 0.003 & 0.219 & 0.053 & 0.220 \\
    \textsc{specter} [T-1] & 0.318 & 0.139 & 0.277 & 0.322 & 0.159 & 0.276 \\
    \textsc{specter} [T-2] & 0.211 & 0.035 & 0.219 & 0.211 & 0.0345 & 0.220 \\
    \textsc{specter} [T-3] & 0.174 & 0.010 & 0.191 & 0.175 & 0.011 & 0.191 \\
    \textsc{t5} & 0.687 & 0.272 & 0.420 & 0.680 & 0.314 & 0.400 \\
    Qwen2.5 & 0.843 & 0.490 & 0.342 & 0.845 & 0.499 & 0.323 \\
    \bottomrule
\end{tabular}
    \vspace{.1in}
    \caption{Comparison of the average correctness score \( S_c \) and the average missing score \( S_m \) for different alignment methods under varying threshold settings. W denotes results obtained using the original transcript before correction. \textbf{T-1}, \textbf{T-2}, and \textbf{T-3} represent different threshold values for textual and visual regions: 
    \textbf{T-1:} Textual region threshold = 0.8, Visual region threshold = 0.6,
    \textbf{T-2:} Textual region threshold = 0.7, Visual region threshold = 0.6
    \textbf{T-3:} Textual region threshold = 0.6, Visual region threshold = 0.6}
    \label{tab:method_scores}
    \vspace{-0.5cm}
\end{table}

\subsubsection{Results of \textsc{asr} Post-Correction with \textsc{ocr} Guidance:}
Table~\ref{tab:ASR-Results} presents a comparison of the \textsc{asr} performance before and after correction. The evaluation was conducted on a subset of the dataset, consisting of 96 samples. Notably, the corrected \textsc{asr} exhibits an increase in Character Error Rate (\textsc{cer}) and Word Error Rate (\textsc{wer}) compared to the original version. This increase is attributed to the \textsc{ocr}-guided correction process, which enforces alignment with the extracted text from slide images. A closer examination of Fig.~\ref{fig:ASR-Post-Cor}, particularly the last two examples, illustrates the nature of these corrections. In the third example, the corrected \textsc{asr} transcribes the number as 91,125 instead of 91,000. This discrepancy arises because the speaker provided an approximate figure during the presentation, whereas the \textsc{ocr} correction enforces the exact value present in the slide. Similarly, in the fourth example, the word "data" is replaced with "datasets" due to \textsc{ocr} guidance, aligning the \textsc{asr} output with the textual content extracted from the slides. Despite such variations, these corrections are beneficial for our task. The adjustments ensure better alignment with the visual content of the slides, aiding in the accurate localization of relevant information. Furthermore, as seen in the first three examples of Fig.~\ref{fig:ASR-Post-Cor}, \textsc{ocr}-guided correction enhances the accuracy of technical terms and key concepts, thereby improving the overall quality of the transcription in a technical presentation setting.

\begin{figure}[!htbp]
    \centering
    \begin{minipage}{0.45\textwidth}
        \centering
        \vspace{0.1in}
        \begin{tabular}{lccc}
            \toprule
            \textbf{Category} & \textbf{A} & \textbf{B} & \textbf{C} \\
            \midrule
            Magnifying/emphasizing & 14.70 & 61.76 & 22.05 \\
            Bounding box & 83.82 & 8.82 & 4.41 \\
            Shading & 57.35 & 10.29 & 32.35 \\
            Hiding the background & 38.23 & 14.07 & 47.05 \\
            \midrule
            Total & 49.26 & 24.26 & 26.47 \\
            \bottomrule
        \end{tabular}
        \caption{User preference rating across different highlighting methods (all values in percentage).}
        \label{tab:clip_summary}
    \end{minipage}%
    \hfill
    \begin{minipage}{0.45\textwidth}
        \centering
        \includegraphics[height=0.55\textwidth, width=\textwidth]{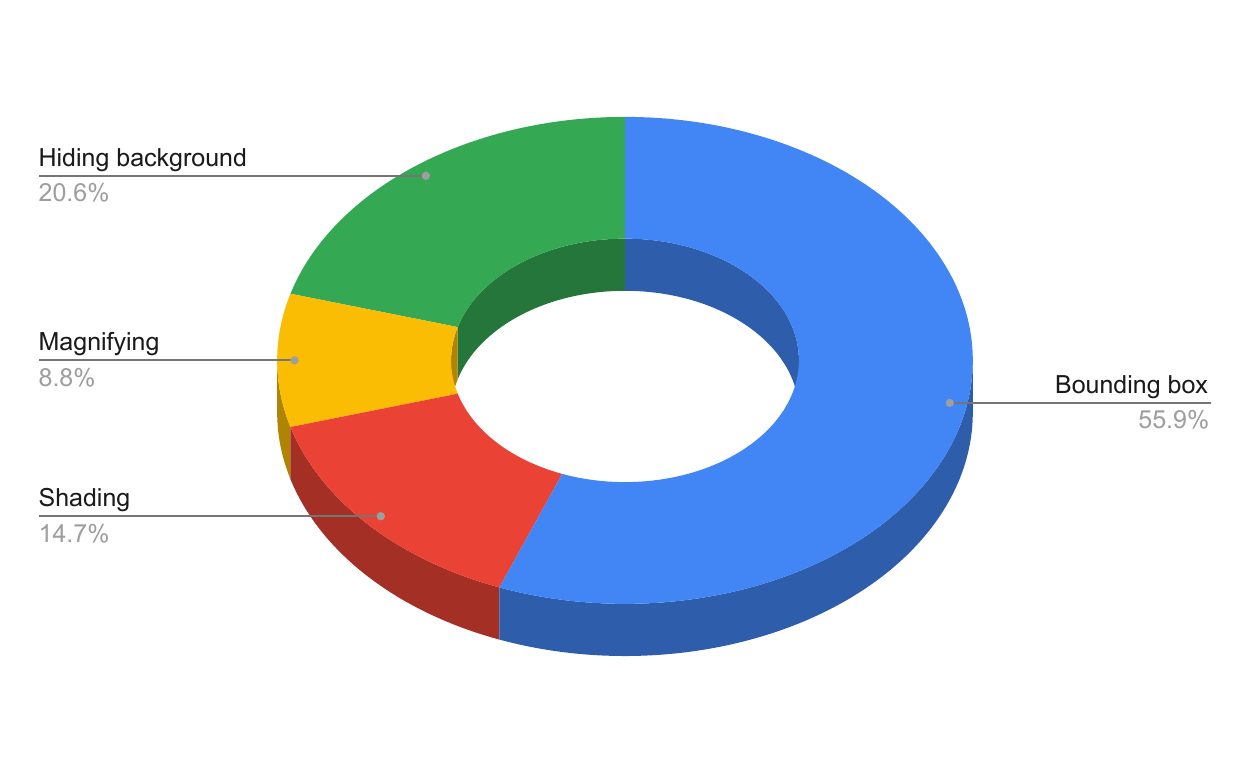}
        \caption{Distribution of user preferences across four different highlighting methods.}
        \label{fig:clip_pie}
    \end{minipage}
\end{figure}

\section{User Study}

We conducted a user study with 33 participants (Graduate, MS, PhD, and those pursuing MS/PhD) to evaluate four highlighting techniques. Each participant viewed six video clips (avg. 35.03s), sampled from the middle of presentations to simulate joining midway. After each clip, participants chose one of the following:
A) The highlighting was useful — I preferred having the highlighting.
B) I didn’t mind whether the highlighting was present or not.
C) I did not prefer the highlighting — I would rather watch the video without it.
At the end, they indicated their most preferred technique (Figure~\ref{fig:enter-label}) and explained why. The results (Figure~\ref{fig:clip_pie}) showed bounding boxes were most preferred. Users found them effective for emphasizing key content while preserving full-slide visibility. Shading drew attention, but some noted it reduced readability by dimming the background. Magnification was least favored—while helpful for small text, it often obscured context and made focus areas unclear. Hiding was generally disliked, as users preferred being able to glance at surrounding content, though some appreciated its ability to reduce distractions. As summarized in Table~\ref{tab:clip_summary}, about 50\% of participants preferred highlighting, especially when joining midway. Highlighting helped users catch up and stay focused, though its effectiveness depended on balancing emphasis with slide visibility.

\section{Conclusion}
We explore aligning speech with slide content using \textsc{ocr} and \textsc{asr} matching, focusing on identifying where to direct attention during a presentation. We evaluated performance using the Correctness Score ($S_c$), Missing Score ($S_m$), and F1 score. We further explored string matching, semantic embeddings, and \textsc{llm}-based approaches for deeper alignment. Future directions include leveraging \textsc{vlm}s for spatial and non-verbal references, and enabling real-time alignment in live scenarios.

\begin{credits}
\section{\ackname} 
This work is supported by the MeitY Government of India, through the NLTM
Bhashini \url{https://bhashini.gov.in/} project.

\end{credits}
%
%
%
\bibliographystyle{splncs04}
\bibliography{main}

\end{document}